\PassOptionsToPackage{usenames}{color}
\pdfoutput=1 
\documentclass[11pt,a4paper]{article}
\usepackage[hyperref]{style/acl2018}

\usepackage{graphicx} 
\usepackage{url} 

\aclfinalcopy 
\def\aclpaperid{1158} 

\usepackage{times}
\usepackage{natbib}

\usepackage{relsize} 

\usepackage{microtype}
\usepackage{ dsfont }
\usepackage[boxed]{algorithm2e}

\usepackage[small,bf,skip=5pt]{caption}
\usepackage{sidecap} 
\usepackage{rotating}	

\usepackage{paralist}

\usepackage{titlesec}
\titleformat*{\subparagraph}{\itshape}
\titlespacing{\subparagraph}{%
  1em}{
  0pt}{
  1em}

\usepackage{lingmacros}

\usepackage[shortlabels]{enumitem} 
\setitemize{noitemsep,topsep=0em} 
\setenumerate{noitemsep,leftmargin=0em,itemindent=13pt,topsep=0em}

\usepackage{xspace}
\usepackage{xparse} 

\usepackage{textcomp}

\usepackage{framed}

\usepackage{listings}

\usepackage{amssymb}	
\usepackage{amsmath}

\usepackage{mathptmx}	
\usepackage[scaled=.8]{beramono}
\usepackage[scaled=.85]{helvet}
\usepackage[T1]{fontenc}
\usepackage[utf8x]{inputenc}

\usepackage{MnSymbol}	

\usepackage{latexsym}

\usepackage{array}
\usepackage{multirow}
\usepackage{booktabs} 
\usepackage{multicol}
\usepackage{footnote}
\newcolumntype{H}{>{\setbox0=\hbox\bgroup}c<{\egroup}@{}} 

\usepackage[usenames]{color}
\usepackage{xcolor}

\usepackage[normalem]{ulem} 
\usepackage{colortbl}
\usepackage{graphicx}
\usepackage{subcaption}
\usepackage{tikz}
\usepackage[edges]{forest}
\usetikzlibrary{arrows,positioning,calc}

\usepackage{nameref}
\usepackage{cleveref}

\crefformat{part}{\S#2#1#3}
\crefformat{chapter}{\S#2#1#3}
\crefformat{section}{\S#2#1#3}
\crefformat{subsection}{\S#2#1#3}
\crefformat{subsubsection}{\S#2#1#3}
\crefformat{paragraph}{\P#2#1#3}
\crefformat{subparagraph}{\P#2#1#3}
\crefmultiformat{section}{\S#2#1#3}{ and~\S#2#1#3}{, \S#2#1#3}{, and~\S#2#1#3}
\crefmultiformat{subsection}{\S#2#1#3}{ and~\S#2#1#3}{, \S#2#1#3}{, and~\S#2#1#3}
\crefmultiformat{subsubsection}{\S#2#1#3}{ and~\S#2#1#3}{, \S#2#1#3}{, and~\S#2#1#3}
\crefmultiformat{paragraph}{\P\P#2#1#3}{ and~#2#1#3}{, #2#1#3}{, and~#2#1#3}
\crefmultiformat{subparagraph}{\P\P#2#1#3}{ and~#2#1#3}{, #2#1#3}{, and~#2#1#3}
\crefrangeformat{section}{\mbox{\S\S#3#1#4--#5#2#6}}
\crefrangeformat{subsection}{\mbox{\S\S#3#1#4--#5#2#6}}
\crefrangeformat{subsubsection}{\mbox{\S\S#3#1#4--#5#2#6}}
\crefrangeformat{paragraph}{\mbox{\P\P#3#1#4--#5#2#6}}
\crefrangeformat{subparagraph}{\mbox{\P\P#3#1#4--#5#2#6}}
\crefname{part}{Part}{Parts}
\Crefname{part}{Part}{Parts}
\crefname{chapter}{ch.}{ch.}
\Crefname{chapter}{Ch.}{Ch.}
\crefname{figure}{figure}{figures}
\crefname{subfigure}{figure}{figures}
\Crefname{subfigure}{Figure}{Figures}
\crefname{appsec}{appendix}{appendices}
\Crefname{appsec}{Appendix}{Appendices}
\crefname{algocf}{algorithm}{algorithms}
\Crefname{algocf}{Algorithm}{Algorithms}
\crefname{enums,enumsi}{example}{examples}
\Crefname{enums,enumsi}{Example}{Examples}
\crefname{}{example}{examples} 
\Crefname{}{Example}{Examples}
\crefformat{enums}{(#2#1#3)}
\crefformat{enumsi}{(#2#1#3)}
\crefrangeformat{enums}{\mbox{(#3#1#4--#5#2#6)}}
\crefrangeformat{enumsi}{\mbox{(#3#1#4--#5#2#6)}}
\crefformat{}{(#2#1#3)}
\crefname{xnumi}{example}{examples} 
\crefname{xnumi}{example}{examples} 
\Crefname{xnumii}{Example}{Examples} 
\Crefname{xnumii}{Example}{Examples} 
\crefformat{xnumi}{(#2#1#3)} 
\crefformat{xnumii}{(#2#1#3)} 
\crefrangeformat{enums}{\mbox{(#3#1#4--#5#2#6)}}
\crefrangeformat{enumsi}{\mbox{(#3#1#4--#5#2#6)}}
\crefrangeformat{xnumi}{\mbox{(#3#1#4--#5#2#6)}} 
\crefrangeformat{xnumii}{\mbox{(#3#1#4--#5#2#6)}} 
\crefmultiformat{enumsi}{(#2#1#3}{, #2#1#3)}{, #2#1#3}{, #2#1#3)}
\crefmultiformat{xnumi}{(#2#1#3}{, #2#1#3)}{, #2#1#3}{, #2#1#3)} 
\crefmultiformat{xnumii}{(#2#1#3}{, #2#1#3)}{, #2#1#3}{, #2#1#3)} 
\crefrangemultiformat{enumsi}{(#3#1#4--#5#2#6}{, #3#1#4--#5#2#6)}{, #3#1#4--#5#2#6}{, #3#1#4--#5#2#6)}
\crefrangemultiformat{xnumi}{(#3#1#4--#5#2#6}{, #3#1#4--#5#2#6)}{, #3#1#4--#5#2#6}{, #3#1#4--#5#2#6)} 
\crefrangemultiformat{xnumii}{(#3#1#4--#5#2#6}{, #3#1#4--#5#2#6)}{, #3#1#4--#5#2#6}{, #3#1#4--#5#2#6)} 

\ifx\creflastconjunction\undefined%
\newcommand{\creflastconjunction}{, and\nobreakspace} 
\else%
\renewcommand{\creflastconjunction}{, and\nobreakspace} 
\fi%

\newcommand*{\Fullref}[1]{\hyperref[{#1}]{\Cref*{#1}: \nameref*{#1}}}
\newcommand*{\fullref}[1]{\hyperref[{#1}]{\cref*{#1}: \nameref{#1}}}

\NewDocumentEnvironment{itmize}{}{\begin{itemize}[noitemsep]}{\end{itemize}}
\NewDocumentEnvironment{enumrate}{}{\begin{enumerate}[noitemsep]}{\end{enumerate}}
\let\Item\item
\renewcommand\enddescription{\endlist\global\let\item\Item}
\NewDocumentEnvironment{describe}{}{\renewcommand\item[1][]{\Item \textbf{##1:} }\begin{itemize}[\null,leftmargin=0em]}{\end{itemize}} 
\NewDocumentEnvironment{edescribe}{}{\renewcommand\item[1][]{\Item \textbf{##1:} }\begin{enumerate}}{\end{enumerate}}

\usepackage{color}
\usepackage{bm}
\definecolor{orange}{rgb}{1,0.5,0}
\definecolor{mdgreen}{rgb}{0.05,0.6,0.05}
\definecolor{mdblue}{rgb}{0,0,0.7}
\definecolor{dkblue}{rgb}{0,0,0.5}
\definecolor{dkgray}{rgb}{0.3,0.3,0.3}
\definecolor{slate}{rgb}{0.25,0.25,0.4}
\definecolor{gray}{rgb}{0.5,0.5,0.5}
\definecolor{ltgray}{rgb}{0.7,0.7,0.7}
\definecolor{purple}{rgb}{0.7,0,1.0}
\definecolor{lavender}{rgb}{0.65,0.55,1.0}

\lstnewenvironment{Python}[1][]
  {\lstset{language=Python,
           morekeywords=as,
           #1}%
  }
  {}
  
\lstnewenvironment{Output}[1][]
  {\lstset{#1}%
  }
  {}

\newcommand{\com}[1]{}

\newcommand{\citeposs}[2][]{\citeauthor{#2}'s (\citeyear[#1]{#2})}

\makeatletter
\renewcommand{\paragraph}{%
  \@startsection{paragraph}{4}%
  {\z@}{.2ex \@plus 1ex \@minus .2ex}{-1em}%
  {\normalfont\normalsize\bfseries}%
}
\makeatother

\newcommand{\w}[1]{\textit{#1}}	
\newcommand{\p}[1]{\textbf{\textsf{#1}}} 
\newcommand{\lbl}[1]{\textsc{#1}} 
\newcommand{\sst}[1]{\lbl{#1}} 
\newcommand{\psst}[1]{\textcolor{mdgreen}{\sst{#1}}} 

\newcommand{\rf}[2]{\psst{#1}$\leadsto$\psst{#2}}

\newcommand{\pex}[1]{\textit{#1}} 

\newcommand{\srsversion}[1]{}

\newcommand{\shortversion}[1]{}

\newcommand{\longversion}[1]{#1} 
\newcommand{\draftnotice}[1]{} 
\newcommand{\nonanonversion}[1]{} 

\usepackage{gb4e} 

\title{Comprehensive Supersense Disambiguation of\\ English Prepositions and Possessives}

\newcommand{\emldisplay}[2]{\texttt{\href{mailto:#1}{#2}}}

\author{
Nathan Schneider\thanks{~~\emldisplay{nathan.schneider@georgetown.edu}{nathan.schneider@georgetown.edu}} \\
	\textsmaller[.5]{Georgetown University} 
     \And
Jena D. Hwang \\
	\textsmaller[.5]{IHMC} 
    \And
Vivek Srikumar \\
	\textsmaller[.5]{University of Utah} 
    \AND
Jakob Prange\\
\bf Austin Blodgett \\
	\textsmaller[.5]{Georgetown University} 
    \And 
Sarah R. Moeller \\
	\textsmaller[.5]{University of Colorado Boulder} 
    \And
Aviram Stern\\
\bf Adi Bitan\\
\bf Omri Abend \\
	\textsmaller[.5]{Hebrew University of Jerusalem} 
}

\date{}

\begin{document}
\maketitle
\begin{abstract}
  Semantic relations are often signaled with prepositional or possessive marking---but extreme polysemy bedevils their analysis and automatic interpretation.
  We introduce a new annotation scheme, corpus, and task for the disambiguation 
  of prepositions and possessives in English.
  Unlike previous approaches, our annotations are comprehensive with respect to types and tokens of these markers; 
  use broadly applicable supersense classes rather than fine-grained dictionary definitions; 
  unite prepositions and possessives under the same class inventory; 
  and distinguish between a marker's \emph{lexical} contribution and the \emph{role} it marks in the context of a predicate or scene.
  Strong interannotator agreement rates, as well as encouraging disambiguation results with established supervised methods, speak to the viability of the scheme and task.
\end{abstract}

\section{Introduction}

Grammar, as per a common metaphor, gives speakers of a language a shared toolbox to construct and deconstruct meaningful and fluent utterances. Being highly analytic, English relies heavily on word order and closed-class function words like prepositions, determiners, and conjunctions. Though function words bear little semantic content, they are nevertheless crucial to the meaning. Consider prepositions: they serve, for example, to convey place and time (\pex{We met \p{at}/\p{in}/\p{outside} the restaurant \p{for}/\p{after} an hour}), to express configurational relationships like quantity, possession, part/whole, and membership (\pex{the coats \p{of} dozens \p{of} children \p{in} the class}), and to indicate semantic roles in argument structure (\pex{Grandma cooked dinner \p{for} the children} vs.~\pex{Grandma cooked the children \p{for} dinner}).
Frequent prepositions like \p{for} are maddeningly polysemous, their interpretation depending especially on the object of the preposition---%
\pex{I rode the bus \p{for} 5 dollars/minutes}---and the governor of the prepositional phrase (PP): \pex{I Ubered\slash asked \p{for} \$5}.
Possessives are similarly ambiguous: \pex{Whistler\p{'s} mother/painting/hat/death}.
Semantic interpretation requires some form of sense disambiguation, 
but arriving at a linguistic representation that is flexible enough to generalize across usages and types, yet simple enough to support reliable annotation, has been a daunting challenge (\cref{sec:background}).

\begin{figure}\centering\small
\begin{exe}\raggedright
\ex I was booked \p{for}/\psst{Duration} 2 nights \p{at}/\psst{Locus} this hotel \p{in}/\psst{Time} Oct 2007 .
\ex I went \p{to}/\psst{Goal} ohm \p{after}/\rf{Explanation}{Time} reading some \p{of}/\rf{Quantity}{Whole} the reviews .
\ex It was very upsetting to see this kind \p{of}/\psst{Species} behavior especially \p{in\_front\_of}/\psst{Locus} \p{my}\slash\rf{SocialRel}{Gestalt} four year\_old .
\end{exe}











\caption{Annotated sentences from our corpus.}
\label{fig:examples}
\end{figure}

This work represents a new attempt to strike that balance.
Building on prior work, we argue for an approach to describing English preposition and possessive semantics with broad coverage. 
Given the semantic overlap between prepositions and possessives (\pex{the hood \p{of} the car} vs.~\pex{the car\p{'s} hood} or \pex{\p{its} hood}), we analyze them using the same inventory of semantic labels.\footnote{Some uses of certain other closed-class markers---intransitive particles, subordinators, infinitive \p{to}---are also included (\cref{sec:lexcat}).}
Our contributions include:
\begin{itemize}
\item a new hierarchical \textbf{inventory} (``SNACS'') of 50~supersense classes, extensively documented in guidelines for English (\cref{sec:scheme});
\item a gold-standard \textbf{corpus} with \emph{comprehensive} annotations: all types and tokens of prepositions and possessives are disambiguated (\cref{sec:corpus}; example sentences appear in \cref{fig:examples});
\item an \textbf{interannotator agreement} study that shows the scheme is reliable and  generalizes across genres---and for the first time demonstrating empirically that the lexical semantics of a preposition can sometimes be detached from the PP's semantic role (\cref{sec:iaa});
\item \textbf{disambiguation} experiments with two supervised classification architectures to establish the difficulty of the task (\cref{sec:disambig}).
\end{itemize}

\section{Background: Disambiguation of Prepositions and Possessives}\label{sec:background}

Studies of preposition semantics in linguistics and cognitive science have generally focused on the domains of space and time \citep[e.g.,][]{herskovits-86,bowerman-01,regier-96,khetarpal-09,xu-10,zwarts-00} or on motivated polysemy structures that cover additional meanings beyond core spatial senses \citep{brugman-81,lakoff-87,tyler-03,lindstromberg-10}.
Possessive constructions can likewise denote a number of semantic relations, and various factors---including semantics---influence whether attributive possession in English will be expressed with \p{of}, or with \p{'s} and possessive pronouns \citep[the `genitive alternation';][]{taylor-96,nikiforidou-91,rosenbach-02,heine-06,wolk-13,shih-15}.

Corpus-based computational work on semantic disambiguation specifically of prepositions and possessives\footnote{Of course, meanings marked by prepositions\slash possessives are to some extent captured in predicate-argument or graph-based meaning representations \citep[e.g.,][]{palmer-05,framenet,oepen-16,amr} and domain-centric representations like TimeML and \mbox{ISO-Space} \citep{timeml,isospace}.} 
falls into two categories: 
the \textbf{lexicographic\slash word sense disambiguation} approach \citep{litkowski-05,litkowski-07,litkowski-14,ye-07,prepnet,dahlmeier-09,tratz-09,hovy-10,hovy-11,tratz-13},
and the \textbf{semantic class} approach \citep[see also \citealp{muller-12} for German]{moldovan-04,badulescu-09,ohara-09,srikumar-11,srikumar-13,schneider-15,schneider-16,hwang-17}.
The lexicographic approach can capture finer-grained meaning distinctions, at a risk of relying upon idiosyncratic and potentially incomplete dictionary definitions. 
The semantic class approach, which we follow here, focuses on commonalities in meaning across multiple lexical items, and aims to generalize more easily to new types and usages.

The most recent class-based approach to prepositions was our initial framework of 75 \textbf{preposition supersenses} arranged in a multiple inheritance taxonomy \citep{schneider-15,schneider-16}. It was based largely on relation\slash role inventories of \citet{srikumar-13} and VerbNet \citep{bonial-11,palmer-17}.
The framework was realized in version 3.0 of our comprehensively annotated corpus, STREUSLE\footnote{\label{fn:streusle}\url{https://github.com/nert-gu/streusle/}}
\citep{schneider-16}.
However, several limitations of our approach became clear to us over time.

First, as pointed out by \citet{hwang-17}, the one-label-per-token assumption in STREUSLE is flawed because it in some cases puts into conflict the semantic role of the PP with respect to a predicate, and the lexical semantics of the preposition itself.
\Citet{hwang-17} suggested a solution, discussed in \cref{sec:construal-analysis}, 
but did not conduct an annotation study or release a corpus to establish its feasibility empirically. 
We address that gap here.

Second, 75~categories is an unwieldy number for both annotators and disambiguation systems. 
Some are quite specialized and extremely rare in STREUSLE~3.0, which causes data sparseness issues for supervised learning. 
In fact, the only published disambiguation system for preposition supersenses collapsed the distinctions to just 12~labels \citep{gonen-16}.
\Citet{hwang-17} remarked that solving the aforementioned problem could remove the need for many of the specialized categories and make the taxonomy more tractable for annotators and systems. 
We substantiate this here, defining a new hierarchy with just 50 categories (SNACS, \cref{sec:scheme}) and providing disambiguation results for the full set of distinctions. 

Finally, given the semantic overlap of possessive case and the preposition \p{of}, 
we saw an opportunity to broaden the application of the scheme 
to include possessives. 
Our reannotated corpus, STREUSLE~4.0, thus has supersense annotations for over 1000 possessive tokens that were not semantically annotated in version 3.0.
We include these in our annotation and disambiguation experiments 
alongside reannotated preposition tokens.

\section{Annotation Scheme}\label{sec:scheme}

\subsection{Lexical Categories of Interest}\label{sec:lexcat} 

Apart from canonical prepositions and possessives, there are many lexically and semantically overlapping closed-class items which are sometimes classified as other parts of speech, such as adverbs, particles, and subordinating conjunctions. 
\emph{The Cambridge Grammar of the English Language} \citep{cgel} argues for an expansive definition of `preposition' that would encompass these other categories. 
As a practical measure, we decided to encourage annotators to focus on the semantics of these functional items rather than their syntax, so we take an inclusive stance.

Another consideration is developing annotation guidelines that can be adapted for other languages.
This includes languages which have \emph{post}positions, \emph{circum}positions, or \emph{in}positions rather than prepositions; the general term for such items is \emph{ad}positions.\longversion{\footnote{In English, \p{ago} is arguably a postposition because it follows rather than precedes its complement: \pex{five minutes \p{ago}}, not *\pex{\p{ago} five minutes}.}}
English possessive marking (via \p{'s} or possessive pronouns like \p{my}) is more generally an example of \emph{case} marking. 
%
%
%
Note that prepositions \cref{ex:gf0,ex:gf1,ex:gf2} differ in word order from possessives \cref{ex:gf3}, though semantically the object of the preposition and the possessive 
nominal pattern together:

\begin{exe}
\ex \label{ex:ground-figure} \begin{xlist}
   \ex \label{ex:gf0} eat \p{in} a restaurant
   \ex \label{ex:gf1} the man \p{in} a blue shirt
   \ex \label{ex:gf2} the wife \p{of} the ambassador
   \ex \label{ex:gf3} the ambassador\p{'s} wife
\end{xlist}
\end{exe}

Cross-linguistically, adpositions and case marking are closely related, and in general both grammatical strategies can express similar kinds of semantic  relations. 
This motivates a common semantic inventory for adpositions and case.

We also cover multiword prepositions (e.g., \p{out\_of}, \p{in\_front\_of}), intransitive particles (\pex{He flew \p{away}}), purpose infinitive clauses (\pex{Open the door \p{to} let in some air}\footnote{\p{To} can be rephrased as \p{in\_order\_to} and have prepositional counterparts like in \pex{Open the door \p{for} some air}.}),  prepositions with clausal complements (\pex{It rained \p{before} the party started}), and idiomatic prepositional phrases (\pex{\p{at}\_large}). 
Our annotation guidelines give further details.

\begin{figure}\centering\small
\newcommand{\hierA}[3]{\textcolor{red}{#1}~~\textsmaller{\textrm{\color{gray}#3}}}
\newcommand{\hierB}[3]{\textcolor{blue}{#1}~~\textsmaller{\textrm{\color{gray}#3}}}
\newcommand{\hierC}[3]{\textcolor{mdgreen}{#1}~~\textsmaller{\textrm{\color{gray}#3}}}
\newcommand{\hierD}[3]{\textcolor{orange}{#1}~~\textsmaller{\textrm{\color{gray}#3}}}
\newenvironment{ggroup}{{}}{{}}
\begin{minipage}{\columnwidth}
\vspace{.2cm}
\begin{multicols}{3}
\begin{ggroup}
  \sffamily\smaller\color{gray}
\begin{forest}
  for tree={%
    folder,
    grow'=0,
    fit=band,
    inner ysep=.75,
  }
  [{\hierA{Circumstance}{76, 63}{77}}
    [{\hierB{Temporal}{0}{0}}
      [{\hierC{Time}{360, 329}{371}}
        [{\hierD{StartTime}{28, 28}{28}}]
        [{\hierD{EndTime}{30, 31}{31}}]
      ]
      [{\hierC{Frequency}{9, 7}{9}}]
      [{\hierC{Duration}{90, 87}{91}}]
      [{\hierC{Interval}{4, 35}{35}}]
    ]
    [{\hierB{Locus}{636, 780}{846}}
      [{\hierC{Source}{77, 189}{189}}]
      [{\hierC{Goal}{234, 378}{419}}]
    ]
    [{\hierB{Path}{26, 44}{49}}
      [{\hierC{Direction}{120, 160}{161}}]
      [{\hierC{Extent}{42, 38}{42}}]
    ]
    [{\hierB{Means}{17, 16}{17}}]
    [{\hierB{Manner}{134, 48}{140}}]
    [{\hierB{Explanation}{121, 108}{123}}
      [{\hierC{Purpose}{301, 396}{401}}]
    ]
  ]
\end{forest}
\columnbreak

\begin{forest}
  for tree={%
    folder,
    grow'=0,
    fit=band,
    inner ysep=.75,
  }
  [{\hierA{Participant}{0}{0}}
    [{\hierB{Causer}{9, 10}{15}}
      [{\hierC{Agent}{158, 37}{170}}
        [{\hierD{Co-Agent}{35, 65}{65}}]
      ]
    ]
    [{\hierB{Theme}{224, 177}{238}}
      [{\hierC{Co-Theme}{14, 7}{14}}]
      [{\hierC{Topic}{213, 289}{296}}]
    ]
    [{\hierB{Stimulus}{123, 0}{123}}]
    [{\hierB{Experiencer}{107, 0}{107}}]
    [{\hierB{Originator}{134, 0}{134}}]
    [{\hierB{Recipient}{122, 0}{122}}]
    [{\hierB{Cost}{48, 30}{48}}]
    [{\hierB{Beneficiary}{93, 76}{110}}]
    [{\hierB{Instrument}{23, 19}{30}}]
  ]
\end{forest}
\columnbreak

\begin{forest}
  for tree={%
    folder,
    grow'=0,
    fit=band,
    inner ysep=.75,
  }
  [{\hierA{Configuration}{0}{0}}
    [{\hierB{Identity}{64, 77}{85}}]
    [{\hierB{Species}{39, 39}{39}}]
    [{\hierB{Gestalt}{165, 699}{709}}
      [{\hierC{Possessor}{381, 489}{492}}]
      [{\hierC{Whole}{142, 173}{250}}]
    ]
    [{\hierB{Characteristic}{133, 66}{140}}
      [{\hierC{Possession}{21, 2}{21}}]
      [{\hierC{PartPortion}{56, 36}{57}}
        [{\hierD{Stuff}{17, 25}{25}}]
      ]
    ]
    [{\hierB{Accompanier}{28, 47}{49}}]
    [{\hierB{InsteadOf}{10, 9}{10}}]
    [{\hierB{ComparisonRef}{176, 184}{215}}]
    [{\hierB{RateUnit}{5, 5}{5}}]
    [{\hierB{Quantity}{191, 84}{191}}
      [{\hierC{Approximator}{76, 73}{76}}]
    ]
    [{\hierB{SocialRel}{240, 0}{240}}
      [{\hierC{OrgRole}{103, 0}{103}}]
    ]
  ]
\end{forest}
\end{ggroup}
\end{multicols}
\end{minipage}
\caption{SNACS hierarchy of 50~supersenses and their token counts in the annotated corpus described in \cref{sec:corpus}.
Counts are of direct uses of labels, excluding uses of subcategories.
Role and function positions are not distinguished (so if a token has different role and function labels, it will count toward two supersense frequencies).}
\label{fig:hier2}
\end{figure}
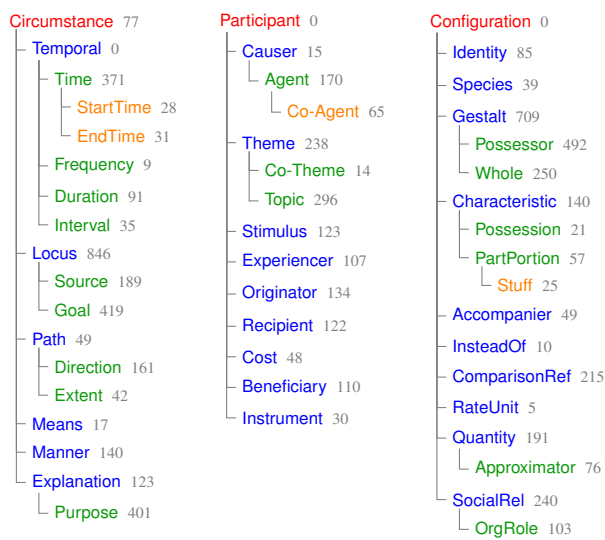

\subsection{The SNACS Hierarchy}\label{sec:hier}

The hierarchy of preposition and possessive supersenses, which we call Semantic Network of Adposition and Case Supersenses (SNACS), is shown in \cref{fig:hier2}. It is simpler than its predecessor---\citeposs{schneider-16} preposition supersense hierarchy---in both size and structural complexity. SNACS has 50 supersenses at 4 levels of depth; the previous hierarchy had 75 supersenses at 7 levels. The top-level categories are the same:
\begin{itemize}
\item \psst{Circumstance}: Circumstantial information, usually 
non-core properties of events (e.g., location, time, means, purpose)
\item \psst{Participant}: Entity playing a role in an event 
\item \psst{Configuration}: Thing, usually an entity or property, involved in a static relationship to some other entity
\end{itemize}
The 3~subtrees loosely parallel adverbial adjuncts, event arguments, and adnominal complements, respectively. The \psst{Participant} and \psst{Circumstance} subtrees primarily reflect semantic relationships prototypical to verbal arguments/adjuncts and were inspired by VerbNet's thematic role hierarchy \citep{palmer-17,bonial-11}. Many \psst{Circumstance} subtypes, like \psst{Locus} (the concrete or abstract location of something), 
can be governed by eventive and non-eventive nominals as well as verbs: \pex{eat \p{in} the restaurant}, \pex{a party \p{in} the restaurant}, \pex{a table \p{in} the restaurant}. \psst{Configuration} mainly encompasses non-spatiotemporal relations holding between entities, such as quantity, possession, and part\slash whole. 
Unlike the previous hierarchy, SNACS does not use multiple inheritance, so there is no overlap between the 3~regions.

The supersenses can be understood as roles in fundamental types of scenes (or schemas) such as: \textsc{location}---\psst{Theme} is located at \psst{Locus}; \textsc{motion}---\psst{Theme} moves from \psst{Source} along \psst{Path} to \psst{Goal}; \textsc{transitive action}---\psst{Agent} acts on \psst{Theme}, perhaps using an \psst{Instrument}; \textsc{possession}---\psst{Possession} belongs to \psst{Possessor}; \textsc{transfer}---\psst{Theme} changes possession from \psst{Originator} to \psst{Recipient}, perhaps with \psst{Cost}; \textsc{perception}---\psst{Experiencer} is mentally affected by \psst{Stimulus}; \textsc{cognition}---\psst{Experiencer} contemplates \psst{Topic}; \textsc{communication}---information (\psst{Topic}) flows from \psst{Originator} to \psst{Recipient}, perhaps via an \psst{Instrument}.
For \psst{Agent}, \psst{Co-Agent}, \psst{Experiencer}, \psst{Originator}, \psst{Recipient}, \psst{Beneficiary}, \psst{Possessor}, and \psst{SocialRel}, the object of the preposition is prototypically animate.

Because prepositions and possessives cover a vast swath of semantic space, 
limiting ourselves to 50~categories means we need to address a great many 
nonprototypical, borderline, and special cases. 
We have done so in a 75-page annotation manual with over 400 example sentences \citep{schneider-18-arxiv}. 

Finally, we note that the Universal Semantic Tagset \citep{abzianidze-17}
defines a cross-linguistic inventory of semantic classes for content and function words. 
SNACS takes a similar approach to prepositions and possessives, 
which in \citeposs{abzianidze-17} specification
are simply tagged \texttt{REL}, 
which does not disambiguate the nature of the relational meaning.
Our categories can thus be understood as refinements to \texttt{REL}.


\subsection{Adopting the Construal Analysis} \label{sec:construal-analysis}


\Citet{hwang-17} have pointed out the perils of teasing apart and generalizing preposition semantics so that each use has a clear supersense label. 
One key challenge they identified is that the preposition itself and the situation as established by the verb may suggest different labels. For instance:
\begin{exe}
\ex \label{ex:work} \begin{xlist}
   \ex\label{ex:work-at-pre} Vernon works \p{at} Grunnings. 
   \ex\label{ex:work-for-pre} Vernon works \p{for} Grunnings. 
\end{xlist}
\end{exe}
The semantics of the \emph{scene} in \cref{ex:work-at-pre,ex:work-for-pre} is the same: 
it is an employment relationship, and the PP contains the employer. 
SNACS has the label \psst{OrgRole} for this purpose.\footnote{\psst{OrgRole} is defined as ``Either a party in a relation between an organization/institution and an individual who has a stable affiliation with that organization, such as membership or a business relationship.''}
At the same time, \p{at} in \cref{ex:work-at-pre} strongly suggests a locational relationship, which would correspond to the label \psst{Locus}; 
consistent with this hypothesis, \pex{Where does Vernon work?} is a perfectly good way 
to ask a question that could be answered by the PP. 
In this example, then, there is overlap between locational meaning and organizational-belonging meaning.
\Cref{ex:work-for-pre} is similar except the \p{for} suggests a notion of \psst{Beneficiary}: the employee is working \emph{on behalf of} the employer. 
Annotators would face a conundrum if forced to pick a single label when multiple ones appear to be relevant.
\citet{schneider-16} handled overlap via multiple inheritance, but entertaining a new label for every possible case of overlap is impractical, as this would result in a proliferation of supersenses.


Instead, \citet{hwang-17} suggest a \textbf{construal analysis} in which the lexical semantic
contribution, or henceforth the \textbf{function}, of the preposition itself may be distinct from the semantic role or relation mediated by the preposition in a given sentence, called the \textbf{scene role}. 
The notion of scene role is a widely accepted idea that underpins the use of semantic or thematic roles: semantics licensed by the governor\footnote{By ``governor'' of the preposition or prepositional phrase, 
we mean the head of the phrase to which the PP attaches in a constituency representation.
In a dependency representation, this would be the head of the preposition itself or of the object of the preposition 
depending on which convention is used for PP headedness: e.g., the preposition heads the PP in CoNLL and Stanford Dependencies whereas the object is the head in Universal Dependencies. 
The governor is most often a verb or noun. Where the PP is a predicate complement (e.g.~\pex{Vernon is \p{with} Grunnings}), there is no governor to specify the nature of the scene, so annotators must rely on world knowledge and context to determine the scene.} of the prepositional phrase dictates its relationship to the prepositional phrase. 
The innovative claim is that, in addition to a preposition's relationship with its head, the prepositional \textit{choice} introduces another layer of meaning or \textbf{construal} that brings additional nuance, creating the difficulty we see in the annotation of \cref{ex:work-at-pre,ex:work-for-pre}.  Construal is notated by \rf{Role}{Function}. 
Thus, \cref{ex:work-at-pre} would be annotated \rf{OrgRole}{Locus} and \cref{ex:work-for-pre} as \rf{OrgRole}{Beneficiary} to expose their common truth-semantic meaning but slightly different portrayals owing to the different prepositions.

Another useful application of the construal analysis is with the verb \w{put}, which can combine with any locative PP to express a destination:
\begin{exe}
   \ex\label{ex:on-put} Put it \p{on}/\p{by}/\p{behind}/\p{on\_top\_of}/\dots~the door. \rf{Goal}{Locus} 
\end{exe}
I.e., the preposition signals a \psst{Locus}, but the door serves as the \psst{Goal} with respect to the scene.
\longversion{This approach also allows for resolution of various semantic phenomena including perceptual scenes (e.g., \pex{I care \p{about} education}, where \p{about} is both the topic of cogitation and perceptual stimulus of caring: \rf{Stimulus}{Topic}), and fictive motion \citep{talmy-96}, where static location is described using motion verbiage (as in \pex{The road runs \p{through} the forest}: \rf{Locus}{Path}).}

Both role and function slots are filled by supersenses from the SNACS hierarchy.
Annotators have the option of using distinct supersenses for the role and function; 
in general it is not a requirement (though we stipulate that certain SNACS supersenses can only be used as the role).
When the same label captures both role and function, 
we do not repeat it: \pex{Vernon lives \p{in}/\psst{Locus} England}.
\Cref{fig:examples} shows some real examples from our corpus.


We apply the construal analysis in SNACS annotation of our corpus to test its feasibility.
It has proved useful not only for prepositions, but also possessives, where the general sense of possession may overlap with other scene relations, like creator\slash initial-possessor (\psst{Originator}): 
\pex{Da Vinci\p{'s}/\rf{Originator}{Possessor} sculptures}.
	


\begin{table}\centering\small
\begin{tabular}{l@{}rrr|r}
                               & \multicolumn{1}{c}{\textbf{Train}} & \multicolumn{1}{c}{\textbf{Dev}}  & \multicolumn{1}{c}{\textbf{Test}} & \multicolumn{1}{c}{\textbf{Total}} \\
\toprule
Documents                      & 347 & 192 & 184 & 723 \\
Sentences                      & 2,723  & 554  & 535  & 3,812 \\
Tokens                         & 44,804 & 5,394 & 5,381 & 55,579 \\
\midrule 
Annotated targets                & 4,522  & 453  & 480  & 5,455 \\
\hspace{1em} Role = function     & 3,101  & 291  & 310  & 3,702 \\
\hspace{1em} P or PP		     & 3,397  & 341  & 366  & 4,104 \\
\hspace{2em} Multiword unit   & 256    & 25   & 24   & 305   \\
\hspace{1em} Infinitive \p{to}   & 201    & 26   & 20   & 247   \\
\hspace{1em} Genitive clitic (\p{'s}) & 52& 6    & 1    & 59    \\
\hspace{1em} Possessive pronoun	 & 872    & 80   & 93   & 1,045 \\
\midrule
Attested SNACS labels            & 47    & 46   & 44  & 47 \\ 
\hspace{1em} Unique scene roles & 46   & 43	  & 41	& 47 \\
\hspace{1em} Unique functions   & 41   & 38	  & 37	& 41 \\
\hspace{1em} Unique pairs 	    & 167  & 79	  & 87	& 177 \\
\hspace{2em} Role = function 	   & 41   & 33	  & 34	& 41 \\
\end{tabular}
\caption{\label{tab:splits}Counts for the data splits used in our experiments. 
}
\end{table}

\section{Annotated Reviews Corpus}\label{sec:corpus}

We applied the SNACS annotation scheme (\cref{sec:scheme}) to prepositions and possessives in the STREUSLE corpus (\cref{sec:background}), a collection of online consumer reviews taken from the English Web Treebank \citep{ewtb}. The sentences from the English Web Treebank also comprise the primary reference treebank for English Universal Dependencies \citep[UD;][]{nivre-16}, and we bundle the UD version~2 syntax alongside our annotations. 
\Cref{tab:splits} shows the total number of tokens present and those that we annotated. Altogether, 5,455 tokens were annotated for scene role and function.


The new hierarchy and annotation guidelines were developed by consensus. The original preposition supersense annotations were placed in a spreadsheet and discussed. While most tokens were unambiguously annotated,  some cases required a new analysis throughout the corpus.
For example, the functions of \p{for} were so broad that they needed to be (manually) clustered 	before mapping clusters onto hierarchy labels. Unusual or rare contexts also presented difficulties. Where the correct supersense remained unclear, specific instructions and examples were included in the guidelines. Possessives were not covered by the original preposition supersense annotations, and thus were annotated from scratch.\footnote{\Citet{blodgett-18} detail the extension of the scheme to possessives.}
\shortversion{Some tokens (marking discourse expressions, infinitives, idioms, etc.)\ were deemed not to be prepositions or possessives evoking semantic relations and were annotated with special labels.}%
\longversion{\textbf{Special labels} were applied to tokens deemed not to be prepositions or possessives evoking semantic relations, 
including uses of the infinitive marker that do not fall within the scope of SNACS (487~tokens: a majority of infinitives) and preposition-initial discourse expressions (e.g.~\pex{\p{after}\_all}) and coordinating conjunctions (\pex{\p{as}\_well\_as}).\footnote{In the corpus, lexical expression tokens appear alongside a \textbf{lexical category} indicating which inventory of supersenses, if any, applies. SNACS-annotated units are those with \textsc{adp} (adposition), \textsc{pp}, \textsc{pron.poss} (possessive pronoun), etc., 
whereas \textsc{disc} (discourse) and \textsc{cconj} expressions do not receive any supersense.
Refer to the STREUSLE README for details.} Other tokens requiring special labels are the opaque possessive slot in a multiword idiom (12~tokens), and tokens where unintelligble, incomplete, marginal, or nonnative usage made it impossible to assign a supersense (48~tokens).}

\begin{table}\centering\small
\begin{tabular}{ccrcr}
Rank & Role & & Function & \\\toprule
1    & \psst{Locus} & 636 & \psst {Locus} & 780 \\
2    & \psst{Possessor} & 381 & \psst{Gestalt} & 699 \\
$\vdots$ & $\vdots$ & & $\vdots$ & \\
last    & \psst{Direction} & 1 & \psst{Possession} & 2 \\
\end{tabular}
\caption{Most and least frequent role and function labels.}
\label{tab:common-labels}
\end{table}

\Cref{tab:common-labels} shows the most and least common labels occurring as scene role and function.
Three labels never appear in the annotated corpus: \psst{Temporal} from the \psst{Circumstance} hierarchy, and \psst{Participant} and \psst{Configuration} which are both the highest supersense in their respective hierarchies. 
While all remaining supersenses are attested as scene roles, there are some that never occur as functions, such as \psst{Originator}, which is most often realized as \psst{Possessor} 
or \psst{Source},
and \psst{Experiencer}.
It is interesting to note that every subtype of \psst{Circumstance} (except \psst{Temporal}) appears as both scene role and function, whereas many of the subtypes of the other two hierarchies are limited to either role or function. This reflects our view that prepositions primarily capture circumstantial notions such as space and time, but have been extended to cover other semantic relations.%
\footnote{All told, 41 supersenses are attested as both role and function for the same token, and there are 136 unique construal combinations where the role differs from the function. Only four supersenses are never found in such a divergent construal: \psst{Explanation}, \psst{Species}, \psst{StartTime}, \psst{RateUnit}. Except for \psst{RateUnit} which occurs only 5 times, their narrow use does not arise because they are rare. \psst{Explanation}, for example, occurs over 100 times, more than many labels which often appear in construal.}


\section{Interannotator Agreement Study}\label{sec:iaa}

Because the online reviews corpus was so central to the development of our guidelines, we sought to estimate the reliability of the annotation scheme on a new corpus in a new genre. We chose Saint-Exup\'{e}ry's 
novella \emph{The Little Prince}, which is readily available in many languages and has been  annotated with semantic representations such as AMR \citep{amr}.
The genre is markedly different from online reviews---it is quite literary, and employs archaic or 
poetic figures of speech. 
It is also a translation from French, 
contributing to the markedness of the language.
This text is therefore a challenge for an annotation scheme based on colloquial contemporary English.
We addressed this issue by running 3 practice rounds of annotation on small passages from \emph{The Little Prince}, both to assess whether the scheme was applicable without major guidelines changes and to prepare the annotators for this genre.
For the final annotation study, we chose chapters 4 and~5, in which
%
242 markables of 52 types were identified heuristically (\cref{sec:pss-id}). 
The types \textit{of}, \textit{to}, \textit{in}, \textit{as}, \textit{from}, and \textit{for}, as well as possessives, occurred at least 10 times.
Annotators had the option to mark units as false positives using special labels (see \cref{sec:corpus}) 
in addition to expressing uncertainty about the unit. 

For the annotation process, we adapted the open source web-based annotation tool UCCAApp \citep{abend-17-uccaapp} to our workflow, 
by extending it with a type-sensitive ranking module for the list of categories presented to the annotators.

\paragraph{Annotators.}
Five annotators (A, B, C, D, E), all authors of this paper, took part in this study. All are computational linguistics researchers with advanced training in linguistics. 
Their involvement in the development of the scheme falls on a spectrum, with annotator A being the most active figure in guidelines development, and annotator E not being involved in developing the guidelines and learning the scheme solely from reading the manual. 
Annotators A, B, and C are native speakers of English, while Annotators D and E are nonnative but highly fluent speakers.
\begin{table}\centering\small
\begin{tabular}{lccc}
	& Labels & Role & Function \\
\toprule
Exact   & 47 & 74.4\% & 81.3\% \\
Depth-3 & 43 & 75.0\% & 81.8\% \\
Depth-2 & 26 & 79.9\% & 87.4\% \\
Depth-1 & 3  & 92.6\% & 93.9\% \\
\end{tabular}
\caption{Interannotator agreement rates (pairwise averages) on \emph{Little Prince} sample (216 tokens) with different levels of hierarchy coarsening according to \cref{fig:hier2} (``Exact'' means no coarsening).
``Labels'' refers to the number of distinct labels that annotators could have provided at that level of coarsening.
Excludes tokens where at least one annotator assigned a non-semantic label.}
\label{tab:iaa}
\end{table}

\paragraph{Results.}
In the \emph{Little Prince} sample,
40 out of 47 possible supersenses 
were applied at least once by some annotator; 36 were applied at least once by a majority of annotators; and 33 were applied at least once by all annotators.
\psst{Approximator}, \psst{Co-Theme}, \psst{Cost}, \psst{InsteadOf}, \psst{Interval}, \psst{RateUnit}, and \psst{Species} were not used by any annotator.

To evaluate interannotator agreement, we excluded 26 tokens for which at least one annotator has assigned a non-semantic label,
considering only the 216 tokens that were identified correctly as SNACS targets and were clear to all annotators.
Despite varying exposure to the scheme, there is no obvious relationship between annotators' backgrounds and their agreement rates.%
\footnote{See \cref{tab:pairwise-iaa} in~\cref{sec:detailed-iaa}
for a more detailed description of the annotators' backgrounds and pairwise IAA results.}

\Cref{tab:iaa} shows the interannotator agreement rates, averaged across all pairs of annotators. 
Average agreement is 74.4\% on the scene role and 81.3\% on the function (row 1).\footnote{Average of pairwise Cohen's $\kappa$ is 0.733 and 0.799 on, respectively, role and function, suggesting strong agreement. However, it is worth noting that annotators selected labels from a ranked list, with the ranking determined by preposition type. The model of chance agreement underlying $\kappa$ does not take the identity of the preposition into account, and thus likely underestimates the probability of chance agreement.}
All annotators agree on the role for 119, and on the function for 139 tokens.
Agreement is higher on the function slot than on the scene role slot, which implies that the former is an easier task than the latter. This is expected considering the definition of construal: the \textit{function} of an adposition is more lexical and less context-dependent, whereas the \textit{role} depends on the context (the scene)
and can be highly idiomatic (\cref{sec:construal-analysis}). 

The supersense hierarchy allows us to analyze agreement at different levels of granularity (rows 2--4 in~\cref{tab:iaa}; see also confusion matrix in supplement). 
Coarser-grained analyses naturally give better agreement, with depth-1 coarsening into only 3~categories. Results show that most confusions are local with respect to the hierarchy. 

\section{Disambiguation Systems}
\label{sec:disambig}

We now describe systems that identify and disambiguate SNACS-annotated prepositions and possessives in two steps.
Target identification heuristics (\cref{sec:pss-id})
first determine which tokens (single-word or multiword) 
should receive a SNACS supersense.  
A supervised classifier then predicts a supersense analysis for each identified target. 
The research objectives are
\begin{inparaenum}[(a)]
\item to study the ability of statistical models to learn roles and functions of prepositions and possessives, and
\item to compare two different modeling strategies (feature-rich and neural), and 
the impact of syntactic parsing.
\end{inparaenum} 


\subsection{Experimental Setup}
\label{sec:experimental-setup}

Our experiments use the reviews corpus described in \cref{sec:corpus}. 
We adopt the official training\slash development\slash test splits of the Universal Dependencies (UD) project; their sizes are presented in \cref{tab:splits}.
All systems are trained on the training set only and evaluated on the test set; 
the development set was used for tuning hyperparameters.
Gold tokenization was used throughout. 
Only targets with a semantic supersense analysis involving labels from \cref{fig:hier2} were included 
in training and evaluation---i.e., tokens with special labels (see \cref{sec:corpus}) were excluded. 

To test the impact of automatic syntactic parsing, 
models in the \textbf{auto syntax} condition were trained and evaluated 
on automatic lemmas, POS tags, and Basic Universal Dependencies (according to the v1 standard) produced by Stanford CoreNLP version~3.8.0~\cite{manning2014stanford}.\footnote{The CoreNLP parser was trained on all 5~genres of the English Web Treebank---i.e., a superset of our training set. Gold syntax follows the UDv2 standard, whereas the classifiers in the auto syntax conditions are trained and tested with UDv1 parses produced by CoreNLP.} 
Named entity tags from the default 12-class CoreNLP model were used in all conditions.




\subsection{Target Identification}\label{sec:pss-id}

\Cref{sec:lexcat} explains that the categories in our scheme apply not only to (transitive) adpositions in a very narrow definition of the term, but also to lexical items that traditionally belong to variety of syntactic classes (such as adverbs and particles), as well as possessive case markers and multiword expressions. 61.2\% of the units annotated in our corpus are adpositions according to gold POS annotation, 20.2\% are possessives, and 18.6\% belong to other POS classes. Furthermore, 14.1\% of tokens labeled as adpositions or possessives are not annotated because they are part of a multiword expression (MWE). It is therefore neither obvious nor trivial to decide which tokens and groups of tokens should be selected as targets for SNACS annotation.

To facilitate both manual annotation and automatic classification, we developed heuristics for identifying annotation targets.
\shortversion{It uses a whitelist and a blacklist of multiword expressions seen in the training data, 
and heuristics based on POS and syntax addressing adpositions, possessives, subordinating conjunctions, adverbs, and infinitivals.}%
\longversion{The algorithm first scans the sentence for known multiword expressions, using a blacklist of non-prepositional MWEs that contain preposition tokens (e.g., \textit{take\_care\_of}) and a whitelist of prepositional MWEs (multiword prepositions like \p{out\_of} and PP idioms like \p{in\_town}). Both lists were constructed from the training data.
From segments unaffected by the MWE heuristics, single-word candidates are identified by matching a high-recall set of parts of speech, then filtered through 5~different heuristics for adpositions, possessives, subordinating conjunctions, adverbs, and infinitivals.
Most of these filters are based on lexical lists learned from the training portion of the STREUSLE corpus, but there are some specific rules for infinitivals that handle \w{for}-subjects (\pex{I opened the door for Steve \p{to} take out the trash}---\p{to}, but not \w{for}, should receive a supersense) and comparative constructions with \textit{too} and \textit{enough} (\pex{too short \p{to} ride}).}

\subsection{Classification}
\label{sec:classifier}
The next step of disambiguation is predicting the role and
function labels. We explore two different modeling strategies.

\paragraph{Feature-rich Model.}
Our first model is based on the features for preposition relation
classification developed by~\citet{srikumar-13}, which were themselves
extended from the preposition sense disambiguation features
of~\citet{hovy-10}. 
We briefly describe the feature set here, and refer the reader to the
original work for further details. At a high level, it
consists of features extracted from selected neighboring words in the
dependency tree (i.e., heuristically identified governor and object) and in the sentence
(previous verb, noun and adjective, and next noun). In addition, all
these features are also conjoined with the lemma of the rightmost
word in the preposition token to capture target-specific
interactions with the labels. The features extracted from each neighboring word are listed in the supplementary material.

Using these features extracted from targets, we trained two multi-class SVM classifiers to
predict the role and function labels using the {\sc liblinear}
library~\cite{fan2008liblinear}.

\paragraph{Neural Model.}
Our second classifier is a multi-layer perceptron (MLP) stacked on top of a BiLSTM.
For every sentence, tokens are first embedded using a concatenation of fixed pre-trained word2vec \cite{mikolov2013distributed} 
embeddings of the word and the lemma, and an internal embedding vector, which is updated during training.\footnote{Word2vec is pre-trained on the Google News corpus. 
Zero vectors are used where vectors are not available.}
Token embeddings are then fed into a 2-layer BiLSTM encoder, yielding a list of token representations.

For each identified target unit $u$, we extract its first token, 
and its governor and  object headword. 
For each of these tokens, we construct a feature vector by concatenating its token representation with embeddings of its
\begin{inparaenum}[(1)]
\item language-specific POS tag, 
\item UD dependency label, and
\item NER label.
\end{inparaenum}
We additionally concatenate embeddings of $u$'s lexical category, 
a syntactic label indicating whether $u$ is predicative\slash stranded\slash subordinating\slash none of these,
and
an indicator of whether either of the two tokens following the unit is capitalized. 
All these embeddings, as well as internal token embedding vectors, are
considered part of the model parameters and are initialized randomly using the Xavier initialization \cite{glorot2010xavier}. 
A \textsc{None} label 
is used when the corresponding feature is not given, both in training and at test time. 
%
%
The concatenated feature vector for $u$ is fed into two separate 2-layered MLPs, followed by a separate softmax layer that yields
the predicted probabilities for the role and function labels. 

We tuned hyperparameters on the development set to maximize $F$-score (see supplementary material).
We used the cross-entropy loss function, optimizing with simple gradient ascent for 80 epochs with minibatches of size 20. 
Inverted dropout was used during training.
The model is implemented with the DyNet library \citep{dynet}. 

The model architecture is largely comparable to that of \citet{gonen-16},
who experimented with a coarsened version of STREUSLE~3.0.
The main difference is their
use of unlabeled multilingual datasets to improve prediction by
exploiting the differences in preposition ambiguities across languages.




\subsection{Results \& Analysis}\label{sec:disambig-results}

\begin{table}[t]\centering\small
\begin{tabular}{lccc}
Syntax & P & R & F \\
\toprule
gold & 88.8 & 89.6 & 89.2 \\
auto & 86.0 & 85.8 & 85.9 \\
\end{tabular}
\caption{Target identification results for disambiguation.}
\label{tab:targetid}
\end{table}

\begin{table*}[]
	\centering\small
	\begin{tabular}{@{}lcccc<{\hspace{5pt}}ccc|ccc|ccc@{}}
                          &        & \multicolumn{3}{c}{\textbf{Gold ID}} & \multicolumn{9}{c}{\textbf{Auto ID}}                                                                                                                     \\
        \cmidrule(r){3-5}\cmidrule(l){6-14}
                          &         & \textit{Role}                        & \textit{Func.} & \textit{Full} & \multicolumn{3}{c}{\textit{Role}} & \multicolumn{3}{c}{\textit{Func.}} & \multicolumn{3}{c}{\textit{Full}}              \\
                          & \textbf{Syntax}       & Acc.                                 & Acc.           & Acc.          & P                                 & R                                  & F    & P    & R    & F    & P    & R    & F    \\
		\midrule
        Most frequent & N/A  & 40.6 & 53.3 & 37.9 & 37.0 & 37.3 & 37.1 & 49.8 & 50.2 & 50.0 & 34.3 & 34.6 & 34.4 \\
        Neural        & gold & 71.7 & 82.5 & 67.5 & 62.0 & 62.5 & 62.2 & 73.1 & 73.8 & 73.4 & 58.7 & 59.2 & 58.9 \\
        Feature-rich  & gold & 73.5 & 81.0 & 70.0 & 62.0 & 62.5 & 62.2 & 70.7 & 71.2 & 71.0 & 59.3 & 59.8 & 59.5 \\
        \midrule
        Neural        & auto & 67.7 & 78.5 & 64.4 & 56.4 & 56.2 & 56.3 & 66.8 & 66.7 & 66.7 & 53.7 & 53.5 & 53.6 \\
        Feature-rich  & auto & 67.9 & 79.4 & 65.2 & 58.2 & 58.1 & 58.2 & 66.8 & 66.7 & 66.7 & 55.7 & 55.6 & 55.7 \\

	\end{tabular}
	\caption{\label{tab:overall} Overall performance of SNACS disambiguation systems on the test set. Results are reported for the role supersense ({\it Role}), the function supersense ({\it Func.}), and their conjunction ({\it Full}). All figures are percentages. 
    \textit{Left:} Accuracies with gold standard target identification (480 targets).
    \textit{Right:} Precision, recall, and $F_1$ with automatic target identification (\cref{sec:pss-id,tab:targetid}).} 
\end{table*}



Following the two-stage disambiguation pipeline (i.e.~target
identification and classification), we separate the evaluation across the
phases. \Cref{tab:targetid} reports the precision, recall, and
$F$-score (P/R/F) of the target identification heuristics. 
\Cref{tab:overall} reports the disambiguation performance of both classifiers with gold (left) and automatic target
identification (right). We evaluate each classifier along three
dimensions---role and function independently, and full (i.e.~both
role and function together). When we have the gold targets, we only
report accuracy because precision and recall are equal. With
automatically identified targets, we report P/R/F for each dimension. Both tables show the impact of syntactic parsing on quality. 
The rest of this section presents analyses of the results along various axes.

\paragraph{Target identification.} 
The identification heuristics described in \cref{sec:pss-id} achieve an $F_1$ score of 89.2\% 
on the test set using gold syntax.\footnote{Our evaluation script counts tokens that received special labels in the gold standard (see \cref{sec:corpus}) as negative examples of SNACS targets, with the exception of the tokens labeled as unintelligible\slash nonnative\slash etc., which are not counted toward or against target ID performance.}
Most  false positives (47/54=87\%) can be ascribed to tokens that are part of a (non-adpositional or larger adpositional) multiword expression.
9 of the 50 false negatives (18\%) are rare multiword expressions not occurring in the training data and there are 7 partially identified ones, which are counted as both false positives and false negatives. 

Automatically generated parse trees
slightly decrease quality (\cref{tab:targetid}). 
Target identification, being
the first step in the pipeline, imposes an
upper bound on disambiguation scores. 
We observe this degradation when we compare the Gold ID and the
Auto ID blocks of \cref{tab:overall}, where 
automatically identified targets decrease $F$-score by about 10
points in all settings.\footnote{A variant of the target ID module, optimized for recall, is used as preprocessing for the agreement study discussed in \cref{sec:iaa}. With this setting, the heuristic achieves an $F_1$ score of 90.2\% (P=85.3\%, R=95.6\%) on the test set.}

\paragraph{Classification.}
Along with the statistical classifier results in \cref{tab:overall}, we also
report performance for the most frequent baseline, which selects
the most frequent role--function label pair given the (gold) lemma
according to the training data.  Note that all learned classifiers,
across all settings, outperform the most frequent baseline for both
role and function prediction. The feature-rich and the neural models
perform roughly equivalently despite the significantly different
modeling strategies. 

\paragraph{Function and scene role performance.} 
Function prediction is consistently more accurate than role prediction,
with roughly a 10-point gap across all systems. This mirrors a
similar effect in the interannotator agreement scores (see
\cref{sec:iaa}), and may be due to the reduced ambiguity of functions compared to roles (as attested by the baseline's higher accuracy for functions than roles), and by the more literal nature
of function labels, as opposed to role labels that often require more
context to determine.

\paragraph{Impact of automatic syntax.}
Automatic syntactic analysis decreases scores by 4 to 7
points, most likely due to parsing errors which affect the
identification of the preposition's object and governor.  
In the auto ID\slash auto syntax condition, 
the worse target ID performance with automatic parses (noted above)
contributes to lower classification scores.


\subsection{Errors \& Confusions}
\label{sec:errors-confusions}

We can use the structure of the SNACS hierarchy
to probe classifier performance. As with the interannotator study, we evaluate the
accuracy of predicted labels when they are coarsened post~hoc by moving up the
hierarchy to a specific depth. \Cref{tab:coarsening-disambig}
shows this for the feature-rich classifier for different
depths, with depth-1 representing the coarsening of the labels into
the 3~root labels. 
Depth-4 (Exact) represents the full results in \cref{tab:overall}. 
These results show that the classifiers often mistake a label for another that is nearby in the hierarchy. 
\begin{table}[t]\centering\small
\begin{tabular}{lccc}
        & Labels    & Role   & Function \\
\toprule
Exact   & 47        & 67.9\% & 79.4\%   \\
Depth-3 & 43        & 67.9\% & 79.6\%   \\
Depth-2 & 26        & 76.2\% & 86.2\%   \\
Depth-1 & 3         & 86.0\% & 93.8\%   \\
\end{tabular}
\caption{Accuracy of the feature-rich model (gold identification and syntax) on the test set (480~tokens) with different levels of hierarchy coarsening of its output.
``Labels'' refers to the number of labels in the training set after coarsening. 
}
\label{tab:coarsening-disambig}
\end{table}
%
%
Examining the most frequent confusions of both models, 
we observe that \psst{Locus} is overpredicted (which makes sense as it is most frequent overall), and \psst{SocialRole}--\psst{OrgRole} and
\psst{Gestalt}--\psst{Possessor}  are often confused (they are close in the hierarchy: one inherits from the other).

\section{Conclusion}

This paper introduced a new approach to comprehensive analysis of the semantics of prepositions and possessives in English, backed by a thoroughly documented hierarchy and annotated corpus.
We found good interannotator agreement and provided initial supervised disambiguation results.
We expect that future work will develop methods to scale the annotation process beyond requiring highly trained experts;
bring this scheme to bear on other languages; 
and investigate the relationship of our scheme to more structured semantic representations, which could lead to more robust models.
Our guidelines, corpus, and software are available at \url{https://github.com/nert-gu/streusle/blob/master/ACL2018.md}.

\section*{Acknowledgments}

We thank Oliver Richardson, whose codebase we adapted for this project; Na-Rae Han, Archna Bhatia, Tim O'Gorman, Ken Litkowski, Bill Croft, and Martha Palmer for helpful discussions and support; and anonymous reviewers for useful feedback.
This research was supported in part by DTRA HDTRA1-16-1-0002/Project \#1553695, by DARPA 15-18-CwC-FP-032, and by grant 2016375 from the United States--Israel Binational Science Foundation (BSF), Jerusalem, Israel.

\bibliographystyle{aclnatbib}
\bibliography{style/pssdisambig}

\newpage

\phantom{xxxxxxx}

\newpage

\appendix


\section{Detailed IAA Analysis}\label{sec:detailed-iaa}

\paragraph{Individual annotators.}
Five annotators took part in this study. All are computational linguistics researchers with advanced training in linguistics. 
Their involvement in the development of the scheme falls on a spectrum:
Annotator~A was the leader of the project and lead author of the guidelines. 
Annotator~B was the second most active figure in guidelines development for an extended period, but took a break of several months in the period when the guidelines were finalized (prior to the pilot study). 
Annotator~C was involved in the later stages of guidelines development.
Annotator~D was involved only at the very end of guidelines development, and primarily learned the scheme from reading the annotation manual. 
Annotator~E was not involved in developing the guidelines and learned the scheme solely from reading the manual (and consulting with the guidelines developers for clarification on a few points).
Annotators A, B, and C are native speakers of English, while Annotators D and E are nonnative but highly fluent speakers.

\Cref{tab:pairwise-iaa} shows that
agreement rates of individual pairs of annotators 
range between 71.8\% and 78.7\% for roles and between 74.1\% and 88\% for functions. 
This is high for a scheme with so many labels to choose from.
Interestingly, there is not an obvious relationship in general between annotators' backgrounds (native language, amount of exposure to the scheme) and their agreement rates.
It is encouraging that Annotators D and E, despite recently learning the scheme from the guidelines, had similar agreement rates to others.

\begin{figure*}[]\small\centering
\includegraphics[width=1.3\textwidth]{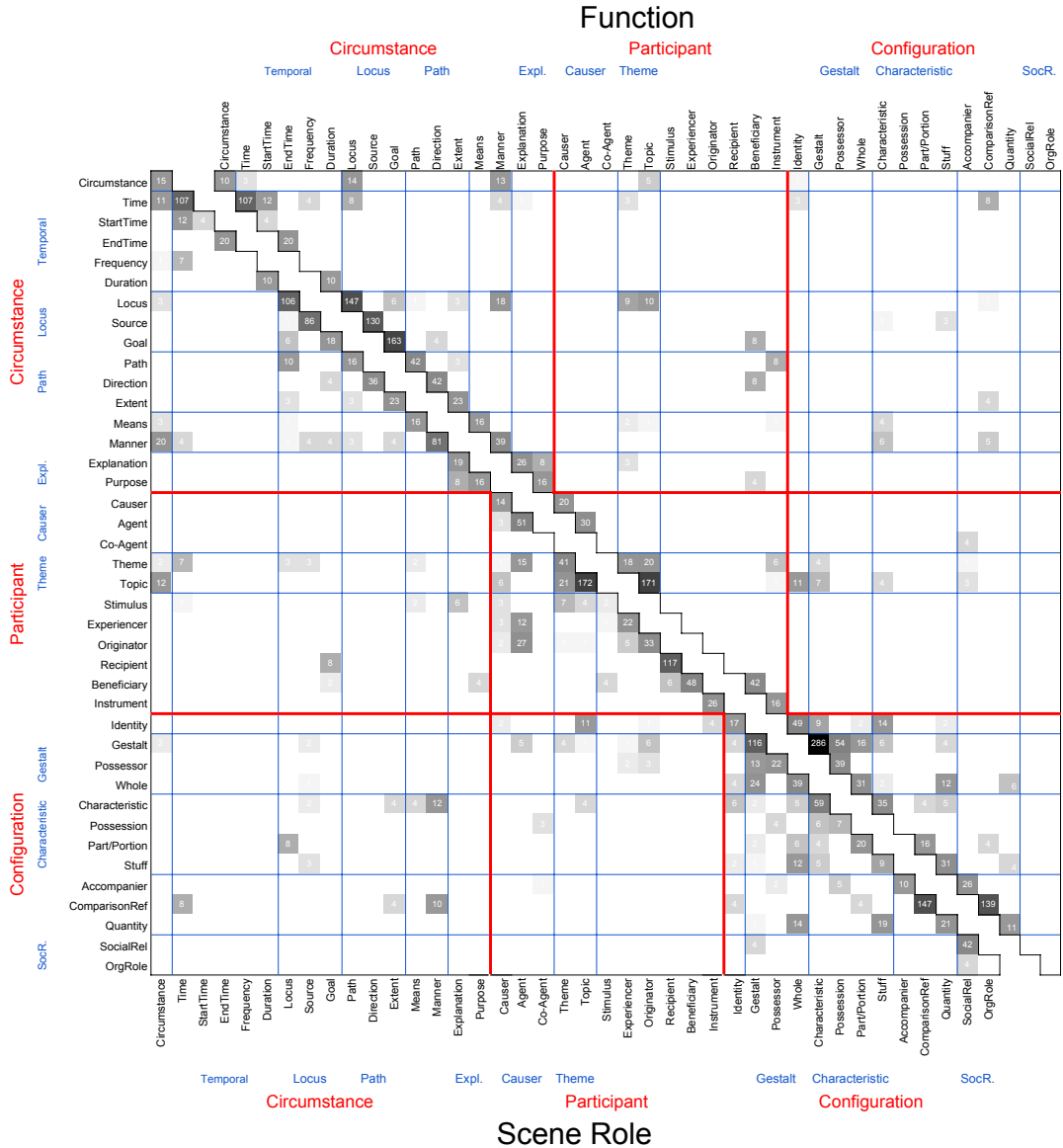}
\caption{Confusion matrices for role (bottom/left) and function (top/right) labels, summed across all annotator pairs.}
\label{fig:confusion}
\end{figure*}

\paragraph{Common confusions.}
In \cref{fig:confusion} we visualize labels confused by annotators in chapters 4 and 5 of \emph{The Little Prince} (\cref{sec:iaa}), summed over all pairs of annotators. The red and blue lines correspond to the local semantic groupings of categories in the hierarchy. Confusions happening within the triangles closest to the diagonal are therefore more expected than confusions farther out in the matrix. 
As discussed in \cref{sec:iaa}, most disagreements actually do fall within these clusters (of varying granularity), indicating the scheme's robustness.

The three most frequently confused scene roles are \psst{Agent}/\psst{Originator} (\textit{\textbf{his} report}, under \psst{Participant}), \psst{Gestalt}/\psst{Whole} (\textit{the soil \textbf{of} that planet}, \psst{Gestalt} is the parent of \psst{Whole}), and \psst{Theme}/\psst{Topic} (\textit{I am not at all sure \textbf{of} success}, \psst{Theme} is the parent of \psst{Topic}).
The three most frequently confused functions are \psst{Gestalt}/\psst{Possessor} (\textit{\textbf{your} planet}, \psst{Gestalt} is the parent of \psst{Possessor}), \psst{Theme}/\psst{Topic}, and \psst{Locus}/\psst{Manner} (\textit{the astronomer had presented it ... \textbf{in} a great demonstration}, both are children of \psst{Circumstance}).

\begin{table}\centering\small
\begin{tabular}{llcccccc}
  &      & B    & C    & D    & E    & avg  & plr \\
\toprule
\multirow{2}{*}{A} & role & 78.2 & 74.1 & 78.7 & 74.5 & 76.4 & 86.1 \\
                   & fxn  & 81.5 & 84.3 & 88.0 & 81.5 & 83.8 & 90.3 \\\midrule
\multirow{2}{*}{B} & role & 	 & 73.1 & 74.5 & 71.8 & 74.4 & 82.9 \\
                   & fxn  &      & 77.3 & 81.0 & 74.1 & 78.5 & 83.8 \\\midrule
\multirow{2}{*}{C} & role &      &      & 73.6 & 72.7 & 73.4 & 80.1 \\
                   & fxn  &      &      & 83.3 & 80.6 & 81.4 & 88.0 \\\midrule
\multirow{2}{*}{D} & role &      &      &      & 73.1 & 75.0 & 84.7 \\
                   & fxn  &      &      &      & 81.0 & 83.3 & 91.7 \\\midrule
\multirow{2}{*}{E} & role &      &      &      &      & 73.0 & 83.3 \\
                   & fxn  &      &      &      &      & 79.3 & 86.1 \\
\end{tabular}
\caption{Pairwise interannotator agreement rates, each annotator's average agreement rate with others (``avg''), and each annotator's rate of agreeing with the label chosen by the plurality of annotators (``plr''). Tokens for which there is no plurality (6 for both role and function) are included and counted as disagreement for all annotators.
Figures are exact label match percentages.}
\label{tab:pairwise-iaa}
\end{table}

\section{Features of the Feature-rich Model}

For each of the neighboring words of the word or phrase to be classified (as described in \cref{sec:classifier}),  we extracted indicator features for:
\begin{enumerate}[nosep]
\item the lowercased word, capitalization, and universal and extended
  POS tags,
\item the word being present in WordNet,
\item WordNet synsets for the first and all senses,
\item the WordNet lemma and lexicographer file name,
\item part, member, and substance holonyms of the word,
\item Roget thesaurus divisions of the word, if it exists,
\item any named entity label associated with the word,
\item its two and three letter character prefixes and suffixes, and
\item common affixes that produce nouns, verbs, adjectives, spatial or
  temporal words, and gerunds.
\end{enumerate}

\section{Hyperparameters for the Neural Model}

\Cref{tab:hyperparams} presents the hyperparameters used by the neural system, for each of the four settings. 

\begin{table*}[]
	\small
	\centering
	\begin{tabular}{@{}l|cccc@{}}
		\toprule
		Hyperparameter & Auto ID/Auto Prep. & Auto ID/Gold Prep. & Gold ID/Auto Prep. & Gold ID/Gold Prep.\\
		\midrule
		External Word2vec embd. dimension & 300 & 300 & 300 & 300 \\
		Token internal embd. dimension & 50 & 100 & 10 & 10 \\
		Update token Word2vec embd.?  & No & No & No & No \\
		Update lemma Word2vec embd.?  & Yes & Yes & Yes & No \\
		MLP layer dimension  & 80 & 80 & 100 & 100 \\
		MLP activation  & tanh & tanh & relu & relu \\
		BiLSTM hidden layer dimension  & 80 & 100 & 100 & 100 \\
		MLP Dropout Prob.  & 0.32 & 0.31 & 0.37 & 0.42 \\
		LSTM Dropout Prob.  & 0.45 & 0.24 & 0.38 & 0.49 \\
		Learning rate  & 0.15 & 0.15 & 0.15 & 0.15 \\
		Learning rate decay  & 0 & 0 & 10^{-4} & 0 \\
		POS embd. dimension  & 5 & 25 & 25 & 5 \\
		UD dependencies embd. dimension  & 5 & 25 & 10 & 25 \\
		NER  embd. dimension  & 5 & 5 & 10 & 5 \\
		GOVOBJ-CONFIG embd. dimension  & 3 & 3 & 3 & 3 \\
		LEXCAT embd. dimension  & 3 & 3 & 3 & 3 \\
		
		\bottomrule
	\end{tabular}
	\caption{\label{tab:hyperparams}
		Selected hyperparameters of the neural system for each of the four settings. With the exception of the external Word2vec embeddings dimension (which is fixed), the parameters were tuned using random grid search on the development set.
	}
	
\end{table*}

\end{document}